%% file: 0.main.tex
\documentclass{article}

\usepackage[preprint]{neurips_2026}

\usepackage[utf8]{inputenc} 
\usepackage[T1]{fontenc}    
\usepackage{hyperref}       
\usepackage{url}            
\usepackage{booktabs}       
\usepackage{amsfonts}       
\usepackage{nicefrac}       
\usepackage{microtype}      
\usepackage{xcolor}         
\usepackage{colortbl}       
\usepackage{amsmath}
\usepackage{amssymb}
\usepackage{mathtools}
\usepackage{amsthm}
\usepackage{float}
\usepackage{wrapfig}
\usepackage{amssymb}

\usepackage{pifont}
\usepackage{tikz}
\usepackage{graphicx}
\usepackage{bm}
\usepackage{verbatim}
\usepackage{balance}
\usepackage{graphicx}
\usepackage{multirow}
\usepackage{xspace}
\usepackage{threeparttable}
\usepackage{enumitem}
\usepackage{makecell}
\usepackage{bbding}
\usepackage{subfig}
\usepackage{float}
\usepackage{wrapfig}
\usepackage[export]{adjustbox}

\newcommand{\vpara}[1]{\vspace{0.04in}\noindent\textbf{#1}\xspace}

\newcommand{\model}{SAO\xspace}
\usepackage{xcolor}

\definecolor{saoRowGray}{gray}{0.92}

\makeatletter
\@ifundefined{hide}{\NewEnviron{hide}{}}{}
\makeatother

\makeatletter
\renewcommand{\@noticestring}{Under review, Feb 2026.}
\makeatother

\title{Advancing Asynchronous Reinforcement Learning for Language Models}
\title{Single-Rollout Asynchronous Optimization for Agentic Reinforcement Learning}

\author{%
Zhenyu Hou\thanks{Equal Contribution. Work done while ZH and YL interned at Z.AI.} \quad
Yujiang Li\footnotemark[1] \quad
Jie Tang \quad
Yuxiao Dong
\\
\\
Tsinghua University
}

\begin{document}

\maketitle

\input{1.abstract}

\begin{figure}[H]
\centering
\includegraphics[width=0.92\textwidth]{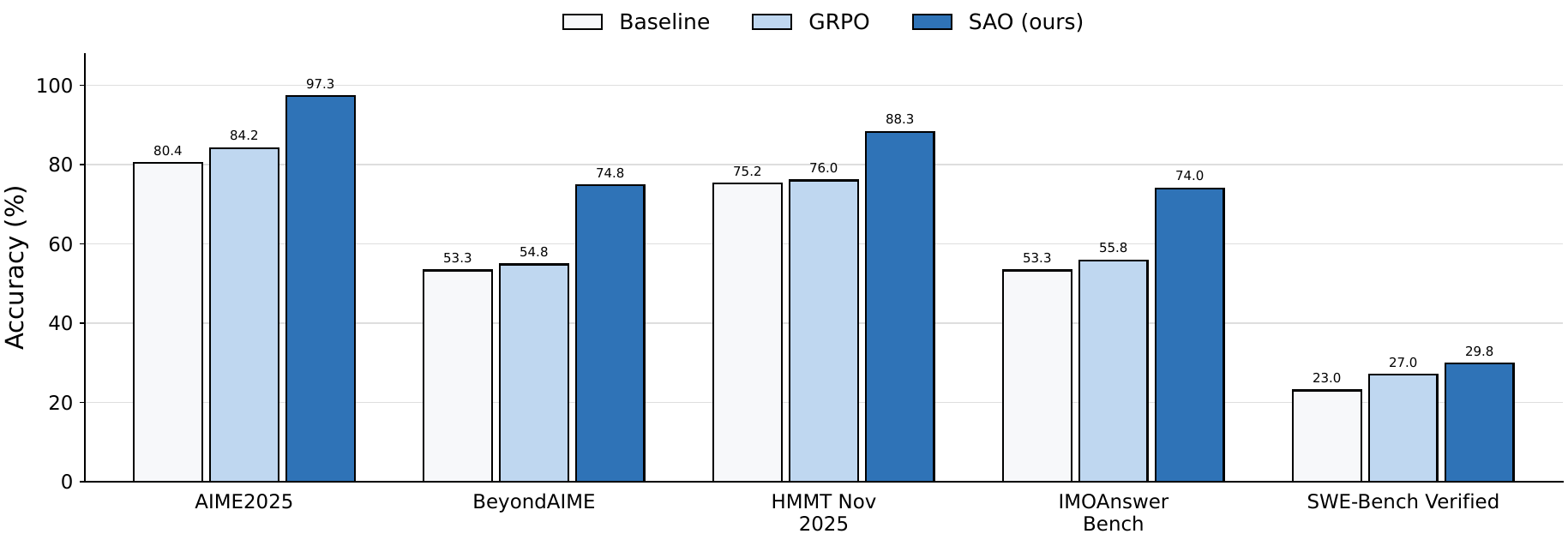}
\vspace{-1mm}
\caption{\textbf{The performance of \model on reasoning and coding benchmarks.} The four reasoning benchmarks are evaluated in a reasoning-with-Python-tool setting, where the baseline is the Qwen3-30B-A3B SFT model; SWE-Bench Verified evaluates coding with the Qwen3-30B-A3B baseline. \model outperforms the corresponding baseline and GRPO across all five benchmarks.}
\label{fig:teaser}
\vspace{-2mm}
\end{figure}

\input{2.introduction}

\input{2.preliminary}
\input{4.method}

\input{5.experiment}
\input{3.related_work}

\input{6.conclusion}

\bibliographystyle{plainnat}
\bibliography{references}

\newpage
\appendix
\input{7.appendix}


\end{document}

%% file: 1.abstract.tex
\begin{abstract}
Reinforcement learning (RL) is becoming increasingly important for post-training large language models (LLMs). 
Previous RL pipelines for LLMs were mostly synchronous and batch-interleaved, which is inefficient for long-horizon agentic tasks. 
Recently, asynchronous RL has emerged as a more efficient alternative by updating the model as rollouts arrive. 
However, existing asynchronous RL systems often emphasize throughput, while leaving training stability and task effectiveness largely underexplored. 
For example, a key challenge is that group-wise sampling in the widely-used GRPO framework does not naturally fit asynchronous agentic training. 
In this paper, we present Single-rollout Asynchronous Optimization (\model{}) to address the stability and off-policy challenges in asynchronous RL. 
To reduce off-policy effects and improve generalization, we replace group-wise sampling with single-rollout sampling, that is, using one rollout per prompt. 
We further improve this single-rollout strategy with practical value-model training designs.
To improve optimization stability, we introduce a strict double-side token-level clipping strategy. 
\model is able to train stably for one thousand steps and consistently outperform GRPO and its variants on agentic coding and reasoning benchmarks, such as SWE-Bench Verified, BeyondAIME, and IMOAnswerBench.  
We also demonstrate that single-rollout RL is particularly effective in a simulated online learning setting, where the model must adapt to changing evolving environments. 
To this end, \model{} is successfully deployed in the agentic RL pipeline for training the open GLM-5.2 model (750B-A40B). 
\end{abstract}


%% file: 2.introduction.tex
\section{Introduction}

Large Language Model (LLM) development is shifting from supervised pre-training toward post-training reinforcement learning (RL). Recent work in Reinforcement Learning has demonstrated that scaling RL compute together with test-time compute is a highly effective way to improve model intelligence \citep{deepseek2024r1, openaio1, cobbe2021training, lightman2023let}.
Most LLM RL pipelines remain synchronous and interleaved: the policy generates a batch of rollouts, and optimization starts only after the entire batch is collected \citep{ouyang2022training, rafailov2024direct}. 

For agentic and coding workloads, rollout lengths are highly variable, so short trajectories finish quickly while long ones become stragglers; as a result, large portions of the GPU cluster idle while waiting for the slowest rollouts \citep{shao2024deepseekv3, kwon2023efficient, yu2022orca}. Asynchronous RL mitigates this \textit{imbalanced generation overhead} by consuming rollouts continuously as they arrive, improving utilization and wall-clock efficiency \citep{mnih2016asynchronous, liang2018rllib, hoffman2020acme}.

However, asynchrony introduces two challenges. 
First, each trajectory can be generated by multiple versions of the old rollout model, which leads to more unpredictable and severe off-policy, and thus harms the training stability. Previous works~\citep{fu2025areal,noukhovitch2024asynchronous} make attempts for asynchronous RL but mainly focus on efficiency optimization rather than effectiveness.
Second, group-wise methods such as GRPO \citep{deepseek2024math, wang2022self} are mismatched to asynchronous training. 
GRPO samples a group of responses for each prompt and uses the group-level average for advantage estimation.
The group-wise sampling induces latency-driven off-policy behavior because the group has to wait for the slower one to finish before fed into training. In addition, group-wise sampling is incompatible with online or complex agentic settings where the environment often provides only a single trajectory feedback per prompt \citep{sutton2018reinforcement, schulman2017proximal, yao2022react, nakano2021webgpt}.


In this paper, we propose Single-rollout Asynchronous Optimization (\model) for  agentic RL. 
It keeps asynchronous RL training  stable and effective under policy lag while preserving the efficiency of asynchrony.
Instead of group-wise sampling, such as GRPO, \model uses single-rollout updates.
To make this setting practical, it also introduces effective value-model training strategies. Our contributions are as follows:

\begin{itemize}[leftmargin=*,itemsep=0pt,parsep=0.2em,topsep=0.3em,partopsep=0.3em]
\item To stabilize training under varied policy lag, we use token-level importance sampling strategy. It directly uses the log-probabilities from the rollout engine and applies stricter double-sided token-level clipping and masking.

\item To reduce off-policy effects, we use one single rollout sampling for each prompt instead of group-wise sampling previously populated by GRPO. 
To further make this setting practical in agentic RL, we improve the value model process. Specifically, we update the critic more frequent than the actor and fine-tune the value model with frozen attention.

\item To handle multi-turn agent trajectories with interleaved environment feedback, we derive a skip-observation token-level GAE estimator. 
It computes advantages across action-to-action boundaries. It also avoids propagating noise through observation tokens that are not generated by the model.

\end{itemize}

We evaluate \model{} on agentic coding and math reasoning benchmarks, including SWE-Bench Verified~\citep{jimenez2023swe}, AIME2025~\citep{balunovic2025matharena}, BeyondAIME~\citep{bytedance_seed_2025_beyondaime}, HMMT\citep{balunovic2025matharena}, and IMOAnswerBench\citep{luong-etal-2025-towards}.
The results demonstrate that our asynchronous RL design can stably train for around one thousand steps and achieves consistently better performance than improved GRPO. In addition, we show that the single-rollout strategy in \model is uniquely suited for simulated online learning, where it can adapt to dynamic environmental changes.

%% file: 2.preliminary.tex
\section{Preliminaries}
In reinforcement learning  for language models, the model is parameterized by $\theta$ as a stochastic policy $\pi_\theta(y|q)$, which generates a response sequence $y = [y_1, \dots, y_{|y|}]$ given a query $q$ from dataset $\mathcal{D}$. RL optimizes $\pi_\theta$ by maximizing a clipped surrogate objective that encourages stable policy updates. Formally, for a given batch of data, the unified optimization target is defined as:

\begin{equation*}
\mathbb{E}\left[ \frac{1}{|y|} \sum_{t=1}^{|y|} \min \left( r_t(\theta) \hat{A}_t, \text{clip}(r_t(\theta), 1-\epsilon, 1+\epsilon) \hat{A}_t \right) \right]
\label{eq:unified_objective}
\end{equation*}

where $r_t(\theta) = \frac{\pi_\theta(y_t \mid q, y_{<t})}{\pi_{\theta_{\text{old}}}(y_t \mid q, y_{<t})}$ is the probability ratio between the current and old policies, $\epsilon$ is the clipping hyperparameter.
The fundamental distinction between PPO~\citep{schulman2017proximal} and GRPO~\citep{shao2024deepseekv3} lies in whether to estimate the advantage function $\hat{A}_t$ and the necessity of auxiliary value networks.

\textbf{Proximal Policy Optimization (PPO).} Standard PPO typically adopts an Actor-Critic architecture, requiring the training of a separate value function (Critic) $V_\phi$, parameterized by $\phi$, to estimate the expected return of the current state. This critic is optimized concurrently with the policy to minimize the value error $\mathcal{L}_\phi^{\text{VF}} = \mathbb{E} [ (V_\phi(q, y_{<t}) - R)^2 ]$, where $R$ denotes the cumulative reward. To balance bias and variance, PPO employs Generalized Advantage Estimation (GAE). The advantage $\hat{A}_t^{\text{GAE}}$ is computed as an exponentially weighted sum of temporal difference errors:

\begin{equation*}
\hat{A}_t^{\text{GAE}} = \sum_{l=0}^{|y|-t-1} (\gamma \lambda)^l \delta_{t+l}
\end{equation*}
where $\delta_t = r_t + \gamma V_\phi(s_{t+1}) - V_\phi(s_t)$.
While effective, this approach necessitates maintaining a copy of the model parameters for the value function, essentially doubling the memory footprint during training and increasing computational overhead.

\begin{hide}
\textbf{Group Relative Policy Optimization (GRPO).} In contrast, GRPO obviates the need for a learned value function and thus reduces memory consumption. Instead of relying on a critic for baseline estimation, GRPO utilizes the properties of a group of sampled responses. For each query $q$, the algorithm samples a group of $G$ outputs $\{y_1, \dots, y_G\}$ from the policy $\pi_{\theta_{\text{old}}}$ and computes their respective rewards $\{R_1, \dots, R_G\}$. The advantage for the $i$-th response is then derived by normalizing its reward against the group:

\begin{equation*}
\hat{A}_{i,t} = \frac{R_i - \mu_R}{\sigma_R}, \quad \text{with} \quad \mu_R = \frac{1}{G}\sum_{j=1}^G R_j
\end{equation*}
where, $\sigma_R = \sqrt{\frac{1}{G}\sum_{j=1}^G (R_j - \mu_R)^2 + \epsilon_{\text{stab}}}$. The group mean $\mu_R$ effectively serves as the baseline, replacing the learned value function $V_\phi$. This formulation renders the advantage $\hat{A}_{i,t}$ constant across all tokens within a specific response $y_i$, assuming sequence-level rewards. 

\end{hide}

%% file: 4.method.tex
\section{Asynchronous Reinforcement Learning with Single Rollout}

\begin{figure}
    \centering
    \includegraphics[width=0.7\linewidth]{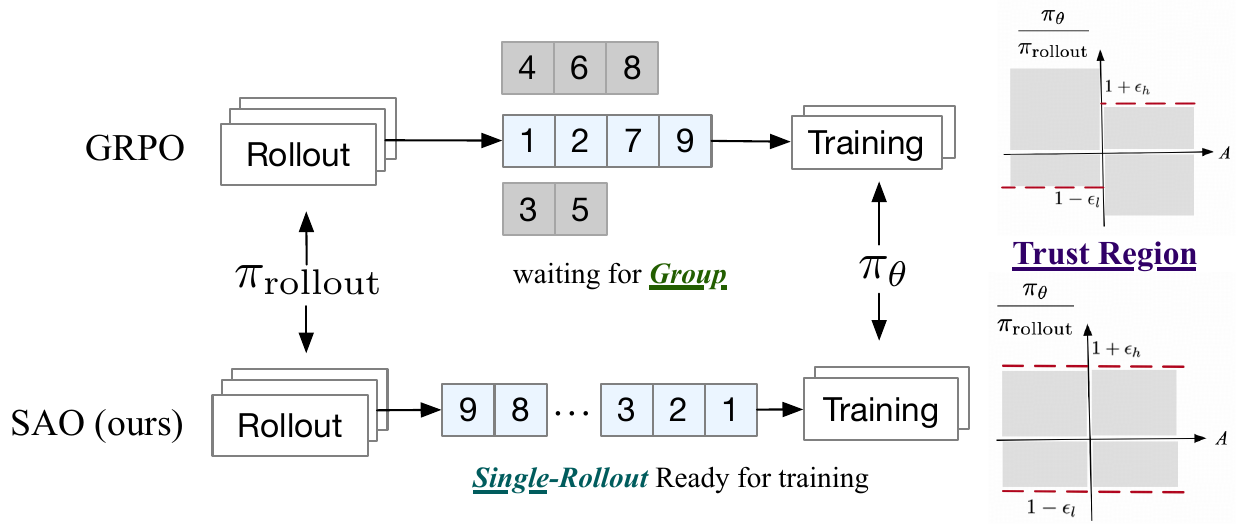}
    \caption{Overview of \model with single rollout design. The numbers denote the generation order of trajectories. For \model, each trajectory becomes available for training immediately upon completion. In contrast, GRPO must wait until all trajectories in a group are generated before training can begin.}
    \label{fig:overview}
\end{figure}

In this section, we introduce \model to tackle training instability and off-policy drift in asynchronous RL training. With a simple token-level clipping strategy and single rollout as an alternative to group-wise sampling, we show that asynchronous RL can be stably scaled to thousands of training steps and achieve significant performance improvements. Figure~\ref{fig:overview} shows the overall design of \model.

\subsection{Stabilizing Asynchronous RL via Direct Double-Sided Importance Sampling (DIS) }

A primary challenge in asynchronous RL is the ``policy lag'' that emerges between rollout models and the training models. In decoupled PPO for LLM, importance sampling is employed to relieve off-policy bias by keeping three distinct models: the current policy $\pi_{\theta}$, the old policy $\pi_{\theta_{\text{old}}}$, and the rollout policy $\pi_{\text{rollout}}$, where $\frac{\pi_{\theta}}{\pi_{\theta_{\text{old}}}}$ is used for staled off-policy correction and $\frac{\pi_{\theta_{\text{old}}}}{\pi_{\text{rollout}}}$ for training-rollout mismatch.
However, as rollout engines may undergo multiple updates during a single trajectory generation in asynchronous RL,
this renders the tracking of exact behavior probabilities $\pi_{\theta_{\text{old}}}$ computationally prohibitive. Otherwise, we have to maintain an extensive history of model checkpoints $\{\pi_{\theta_{\text{old}}^{(1)}}, \dots, \pi_{\theta_{\text{old}}^{(N)}}\}$, which is infeasible in practical implementation.

To resolve this, we propose a simplified yet aggressive token-level importance sampling to clip off-policy tokens.
First, we directly use $\pi_{\text{rollout}}$ as the behavior proxy and $\pi_{\theta}$ for importance sampling, i.e., $r_t(\theta)=\frac{\pi_{\theta}}{\pi_{\text{rollout}}}$, while dropping the inaccurate $\pi_{\theta_{\text{old}}}$. This eliminates the computational overhead of separate old-policy inference by utilizing the log-probabilities generated during the rollout phase.

Second, we employ a double-sided calibration token-level masking strategy. Unlike standard PPO clipping, which clips only selected off-policy tokens with $(A>0, r_t(\theta) > 1+\epsilon_h)$ or $(A<0, r_t(\theta)<1-\epsilon_l)$, we restrict the trust region to the interval $[1-\epsilon_\ell, 1+\epsilon_h]$, while tokens falling outside this range are masked from gradient computation entirely to prevent instabilities arising from extreme policy divergence. This shares similarities with the IcePop mechanism \cite{team2025every}, yet our strategy is simpler by further removing $\pi_{\theta_{\text{old}}}$ while still achieving stable training.

Formally, the optimization objective with token-level clipping can be written as:
\begin{equation}
    L(\theta) = \hat{\mathbb{E}}_t \left[ f(r_t(\theta), \epsilon_l, \epsilon_h) \hat{A}_t \log \pi_{\theta}(a_t|s_t) \right]
\end{equation}
In this formulation, the probability ratio $r_t(\theta)$ is computed directly from the rollout logs to circumvent the need for historical policy tracking:
\begin{equation}
    r_t(\theta) = \exp\left( \log \pi_\theta(a_t|s_t) - \log \pi_{\text{rollout}}(a_t|s_t) \right)
\end{equation}
Stability is further enforced via the calibration function $f(x; \epsilon_\ell, \epsilon_h)$:
\begin{equation}
    f(x; \epsilon_\ell, \epsilon_h) = 
    \begin{cases} 
    x, & \text{if } 1-\epsilon_\ell < x < 1+\epsilon_h \\
    0, & \text{otherwise}
    \end{cases}
\end{equation}

This design circumvents the intensive need to track the historical model ensemble. By utilizing the rollout log-probabilities directly, we accept a controlled degree of off-policy bias in exchange for a substantial reduction in computational complexity and the elimination of errors associated with using a single, potentially stale, ``latest'' old policy model. Empirical results demonstrate that this simplified mechanism enables more aggressive clipping, which effectively regularizes the update steps and yields superior training stability in asynchronous settings.

\subsection{Reducing Off-Policy with Single Rollout}
In asynchronous RL, an inevitable problem is off-policy. Yet current popular group-wise sampling RL algorithms, e.g., GRPO, could introduce more severe off-policy.
Group-wise sampling introduces an ``imbalanced generation'' bias, and the group data has to wait for the ``slowest'' sample to finish before being fed into training.
One promising solution is to replace group-wise sampling with single-rollout, where a sample is immediately fed into training upon generation.

However, single-rollout optimization inherently suffers from high variance in gradient estimation, similar to REINFORCE~\cite{zhang2021sample}. To reduce variance requires a sufficiently good value model. In this part, we focus on simple strategies to optimize value modeling to ultimately boost the policy's performance.  

\vpara{Faster Value Update than Policy.}
We identify that the primary source of instability in single-rollout RL is the interdependence between the policy and the value function. If the value model $V_\phi$ is inaccurate, the advantage estimates $\hat{A}_t$ become noisy, leading to destructive policy updates.
To mitigate this, we implement a Faster Value Update adapted for LLMs. We decouple the optimization frequencies of the policy and the value model. Specifically, for every single gradient update applied to the policy $\pi_{\theta}$, we enforce $K$ updates to the value network $V_{\phi}$ (where $K > 1$).
In our experiments, we set $K=2$. This strategy facilitates the faster adaptation of value estimates to the current policy before they are utilized for advantage computation, thereby reducing the variance.

\vpara{Stabilizing Value Model Training via Parameter Freezing.}
In our pilot experiments, we find the instability of value model training, where the gradient norms of the value model are significantly larger than the corresponding policy model. Further decomposition shows that this instability originates primarily from the Full Attention layers, whereas the Mixture-of-Experts (MoE) layers remain relatively stable.
Based on this observation, we employ a ``Frozen-Attention'' training strategy for the value model. During the RL training, we freeze the parameters of the attention modules in $V_\phi$ and optimize the MoE projections.
We hypothesize that the pre-trained attention weights already possess sufficient semantic capability to attend to relevant tokens. By restricting optimization to the MoE layers, we effectively regularize the value model.

\vpara{Skip-Observation Token-level GAE for Agentic Tasks.}
Agentic tasks present a unique challenge for token-level value estimation due to their trajectory structure: $T = [a_0, o_0, a_1, o_1, \dots]$, where $a_i$ represents model actions and $o_i$ represents environment feedback. Standard Generalized Advantage Estimation (GAE) attempts to calculate the value difference between adjacent tokens. However, the transition from the end of an action $a_{i, \text{end}}$ to the start of an observation $o_{i, \text{start}}$ is discontinuous from the model's perspective, as the model does not generate $o_i$.
Calculating advantage across this boundary introduces noise, as the value model $V(o_{i, \text{start}})$ attempts to predict the value of an external environment state.

To resolve this, we derive a ``Skip-Observation'' GAE. We explicitly modify the Bellman target to bypass environment feedback tokens, linking the value of the current action directly to the value of the subsequent action.
Formally, let $a_{i, N}$ be the last token of action $i$, and $a_{i+1, 0}$ be the first token of the next action. We define the advantage as:

\begin{equation}
    \hat{A}(a_{i, N}) = \delta + \gamma \lambda \hat{A}(a_{i+1, 0})
\end{equation}

where the temporal difference residual $\delta$ is calculated bridging the observation gap:

\begin{equation}
    \delta = r_t + \gamma V(a_{i+1, 0}) - V(a_{i, N})
\end{equation}

This formulation constrains the advantage estimation to rely purely on the model outputs, filtering out the stochasticity of environment feedback. In contrast, some works may consider using a step-level value function and GAE as an alternative to the token-level value; however, we found that a step-level value could lead to suboptimal performance, which will be shown in the experimental part.
We also conduct other advantage designs for agentic traces, and the results can be found in the Appendix. 

\vpara{Scaling Value Pretraining.}
Finally, to support these mechanisms, we find it essential to scale the data used for value model pretraining. Our experiments demonstrate that the ``cold start'' problem in value estimation is a major bottleneck. By significantly increasing the scale of the value pretraining corpus, we provide a robust initialization point that promotes the effectiveness of our single-rollout and TTUR mechanisms from the early stages of training.

%% file: 5.experiment.tex
\section{Experiments}

\begin{table*}[t]
\caption{Experimental Results on math reasoning benchmarks(Accuracy \%).}
\centering
\setlength{\tabcolsep}{4pt}
\begin{tabular}{lcccc}
\toprule
Model & AIME2025 & BeyondAIME & HMMT Nov 2025 & IMOAnswerBench  \\
\midrule
Claude-Sonnet-4.5 & 87.0 & 62.0 & 81.7 & 65.8 \\
GPT-5 High & 94.6 & 74.0 & 89.2 & 76.0 \\
GLM-4.7 & 95.7 & - & 93.5 &  82.0\\
\midrule
Qwen3-30B-A3B  \\
\quad w/ python & 14.6 & 10.5 & 17.3 & 7.8 \\
\quad w/o python & 85.0 & 63.0 & 76.7 & 55.3 \\
\quad SFT (w/ python) & 80.4 & 53.3 & 75.2 & 53.3 \\
\quad SFT (w/o python) & 14.6 & 46.8 & 17.3 & 42.0 \\
\quad GRPO (w/ python) & 84.2 & 54.8 & 76.0 &  55.8\\
  
 \midrule
Qwen3-30B-A3B \\
\rowcolor{saoRowGray}
\quad  \model (ours) & \textbf{97.3} & \textbf{74.8} & \textbf{88.3} & \textbf{74.0} \\
\quad  - \textit{SAO (w/ DIS only)} & 94.2 & 71.5 & 86.7 & 71.3\\
\quad - \textit{GRPO (+ DIS)} & 93.5 & 70.8 & 84.0 & 70.0 \\

\bottomrule
\end{tabular}
\label{tab:main_results}
\end{table*}

\begin{table*}[t]
    \centering
    \caption{Experimental Results  on SWE-Bench Verified (Accuracy \%).}
    \begin{tabular}{l|c}
    \toprule
     Model    & Accuracy (\%) \\
    \midrule
    Qwen3-30B-A3B    &  23.0 \\
    + GRPO (w/ DIS)     & 27.0 \\
\rowcolor{saoRowGray}
    + \model (ours) & 29.8 \\
    \bottomrule
    \end{tabular}
    \label{tab:swe_bench}
\end{table*}

\begin{figure*}[t]
\centering
  \includegraphics[width=0.32\textwidth]{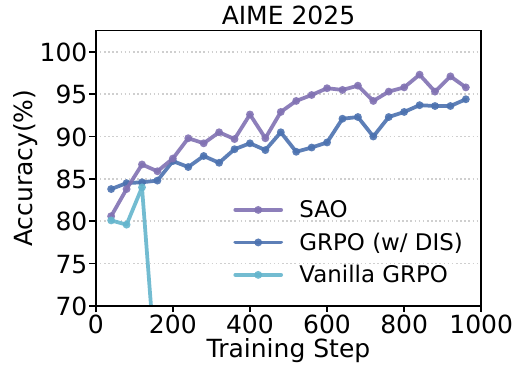}
\hfill
  \includegraphics[width=0.32\textwidth]{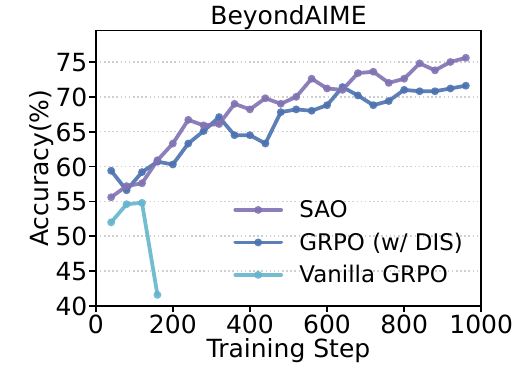}
\hfill
  \includegraphics[width=0.32\textwidth]{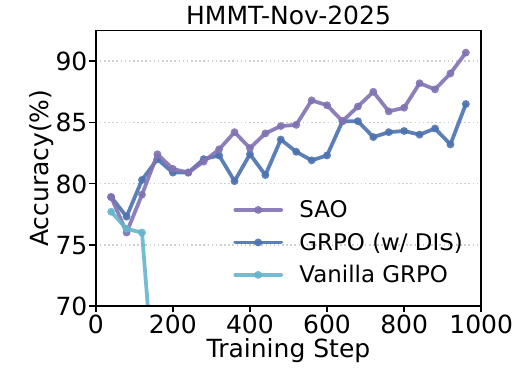}
\caption{Performance comparison between \model and GRPO (w/ DIS) during training. It can be observed that \model almost consistently outperforms the optimized GRPO during the training process on different benchmarks. 
}
\label{fig:main_results}
\vspace{-4mm}
\end{figure*}

\subsection{Experimental Setup}
\vpara{Training Details.}
For math reasoning with Python, we finetune Qwen3-30B-A3B-Thinking-2507~\citep{yang2025qwen3} for 3 epochs on Tool-Integrated Reasoning (TIR) data produced by GPT-OSS-120B~\citep{openai2025gptoss120bgptoss20bmodel} and use the finetuned model to initialize the policy and value model. TIR requires the model to interleave natural-language math reasoning with Python tool calls.

For RL of agentic reasoning, we employ a batch size of 128, a group size of 1, and a max-length of 128k tokens. The policy is optimized with a learning rate of $1 \times 10^{-6}$, with a token clipping of $\epsilon_{\text{low}}=0.3$, $\epsilon_{\text{high}}=5.0$. 
We adopt a length-adaptive GAE~\citep{yue2025vapo} with $\lambda_{\text{policy}} = 1 - \frac{1}{\alpha l}$ and $\alpha = 1.5$. The value model is trained with a learning rate of $5 \times 10^{-6}$, $\lambda_{\text{critic}} = 1$, and a 10-step warmup period. We set the $K=2$ for faster value update for the value model, performing two value model updates per batch. For GRPO variants, each training batch contains 16 prompts with 8 rollout samples per prompt, yielding the same batch size of 128.
For the RL of coding agent, we directly use Qwen3-30B-A3B-Thinking-2507 for training and keep almost all the hyperparameters the same as TIR, except for $\epsilon_{\text{low}}=0.8$ and $\epsilon_{\text{high}}=3.0$. For SWE-Bench Verified, we use OpenHands as the scaffold, with a maximum of 300 interaction turns and a 128k-token context budget.

\vpara{Evaluation.} We evaluate \model on four math reasoning benchmarks including AIME2025, BeyondAIME\citep{bytedance_seed_2025_beyondaime}, 
HMMT Nov 2025\citep{balunovic2025matharena} and IMOAnswerBench\citep{luong-etal-2025-towards}, reporting Pass@1 accuracy. All evaluations use top-$p=1.0$, temperature $1.0$, and a maximum generation length of 128k tokens. Math-reasoning evaluations allow up to 50 turns to support extensive reasoning and tool calls, while SWE-Bench Verified evaluations allow up to 300 OpenHands interaction turns. To reduce variance, we report the mean performance across 16 evaluation runs for AIME2025 / HMMT / IMOAnswerBench and 4 runs for BeyondAIME.


\subsection{Main Results}
Tables \ref{tab:main_results} and \ref{tab:swe_bench} summarize the performance of baselines and different training strategies. GRPO denotes the standard GRPO with \textit{clip-higher} implementation~\cite{yue2025vapo}, which keeps the latest old policy for importance sampling. GRPO (w/ DIS) denotes using the proposed DIS strategy for GRPO.





As shown in Tables \ref{tab:main_results} and ~\ref{tab:swe_bench}, \model consistently outperforms all baselines on both agentic reasoning and coding benchmarks. Standard GRPO
suffers from a performance collapse at approximately 160
training steps
The scores reported for these models represent their final valid performance before collapsing.
Figure \ref{fig:main_results} illustrates the evaluation performance across training steps of \model compared to vanilla GRPO and GRPO (w/ DIS). Vanilla GRPO tends to quickly collapse, while GRPO with DIS can achieve stable training, demonstrating the effectiveness of DIS. 
In addition, \model and GRPO (w/ DIS) exhibit comparable performance in the initial stage; a distinct performance divergence occurs after approximately 400 training steps, demonstrating the effectiveness and stability of \model.

\subsection{Ablation Studies}
We conduct extensive ablation studies to evaluate the impact of various training configurations on the performance of \model. The results are shown in Table~\ref{tab:ablations}.

\begin{itemize}[leftmargin=*,itemsep=0pt,parsep=0.2em,topsep=0.3em,partopsep=0.3em]
\item \textit{Effects of faster value update}: To ablate the effects of faster value update than the policy model, we conduct experiments where the value model is updated only once per batch (critic-train-epoch=1), as opposed to the two updates per batch employed in \model. 
\item \textit{Full vs. frozen-attention value model.}: To evaluate the impact of attention-freezing during RL, this variant performs full-parameter updates on the value model. 

\item \textit{Vanilla VAPO and single-rollout with Running-mean baseline.} Standard VAPO\citep{yue2025vapo} with length-adaptive GAE and a value-based RL baseline. Besides, a single-rollout baseline that maintains a sliding window of the 8 most recent rewards for each prompt, using their mean as a baseline for advantage estimation to provide a simple alternative to parametric value models.

\end{itemize}

As shown in Table \ref{tab:ablations}, all examined variants exhibit a performance decline relative to the proposed \model, validating the necessity of each design choice. Table~\ref{tab:value_training_ablations} further summarizes the value-training strategy and critic-update settings behind the main value-model ablations. Regarding update frequency, the results indicate that a single update is insufficient for the critic to accurately track rapid policy shifts, leading to less reliable baseline estimations. The full-parameter value-training variant further suggests that frozen-attention updates help regularize critic optimization in complex reasoning tasks.
In addition, the RL with running-mean reward achieves decent performance, but still lags far from our \model, demonstrating the advantage and necessity of a well-trained value model for RL. As for the vanilla VAPO, the training also quickly collapses during training, similar to the vanilla GRPO. 

\begin{figure*}[htbp]
\centering
\subfloat[]{
  \includegraphics[width=0.32\textwidth]{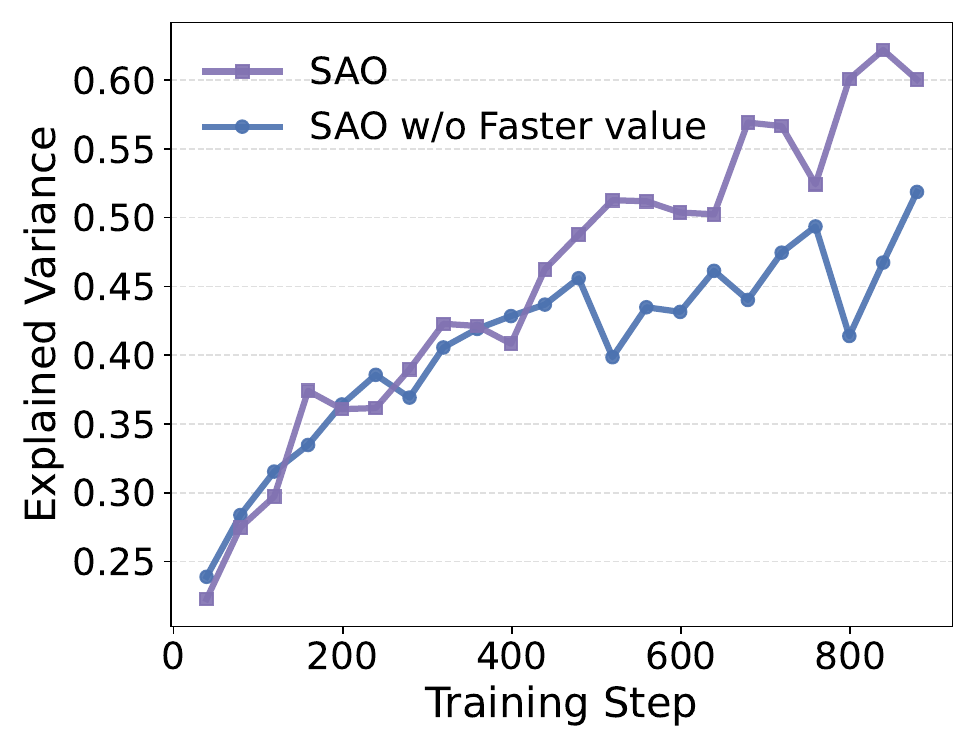}
}
\hfill
\subfloat[]{
  \includegraphics[width=0.32\textwidth]{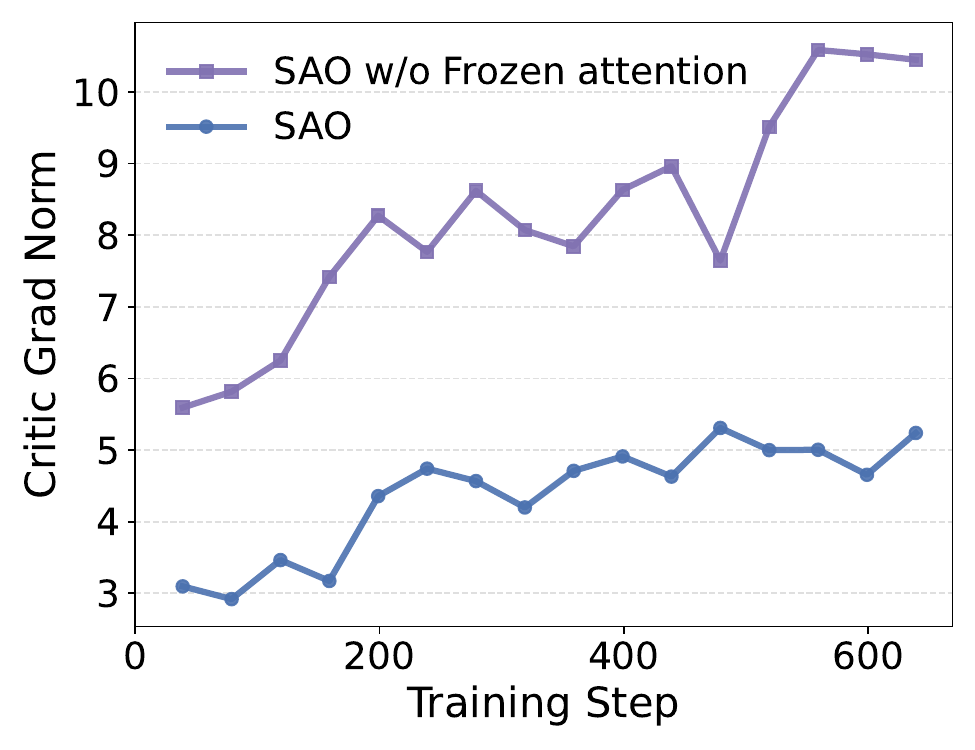}
}
\hfill
\subfloat[]{
  \includegraphics[width=0.32\textwidth]{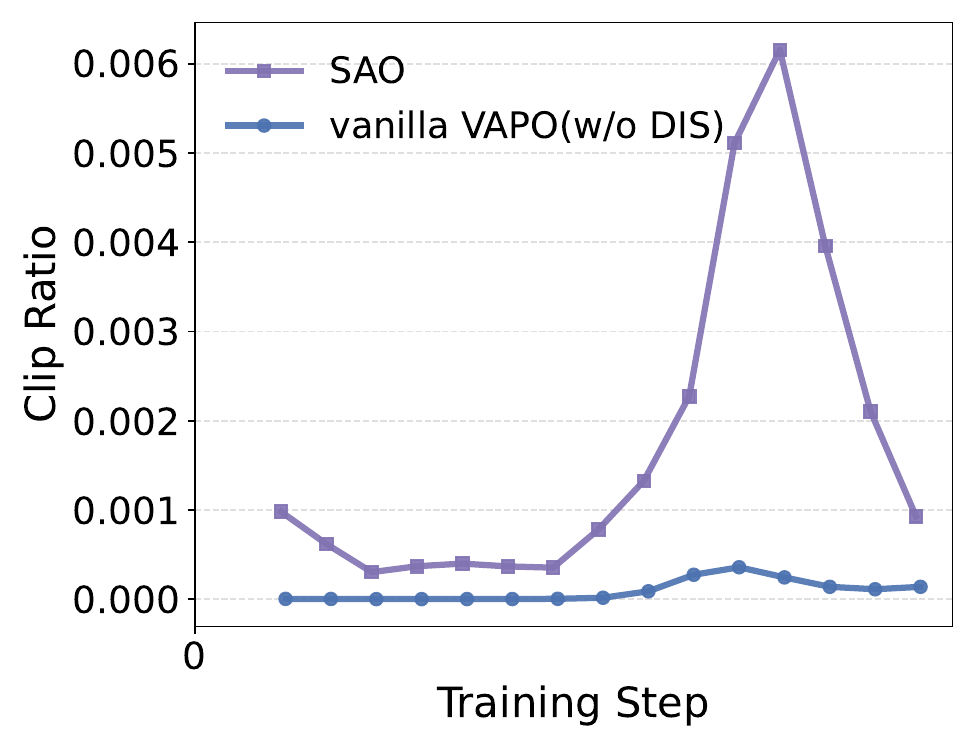}
}
\caption{
Training dynamics of asynchronous single-rollout RL.
(a) Explained Variance for \model and a single-critic-update baseline.
(b) Critic gradient norm during value training under full-parameter optimization and frozen-attention optimization used in \model.
(c) Token-level clip ratio during training for \model with the proposed DIS and the VAPO baseline.
}
\label{fig:training_dynamics}
\end{figure*}

\subsection{Training Dynamics}
We analyze the training dynamics of \model to understand how it facilitates training stability.

\vpara{Effects of Faster Value Update.} Figure \ref{fig:training_dynamics}(a) illustrates Explained Variance comparison between \model and the single-critic-update baseline during training. Explained Variance assesses the alignment between the predicted values $V(s)$ and the ground-truth returns $R$, defined as $EV = 1 - \frac{\text{Var}(R - V(s))}{\text{Var}(R)}$. 
\model demonstrates significantly higher explained variance after approximately 400 training steps, indicating faster value convergence and better alignment with policy distribution.

\vpara{Gradient of Critic Models.} We examine the impact of freezing attention parameters in value training. As shown in Figure \ref{fig:training_dynamics}(b), full-parameter value training exhibits significantly larger critic gradient norms, suggesting unstable optimization dynamics. In contrast, the frozen-attention strategy maintains lower and smoother gradient norms, implying improved numerical stability.

\vpara{Clipped Tokens.} Figure \ref{fig:training_dynamics}(c) monitors the token-level clip ratio of \model applying our proposed DIS strategy and the standard VAPO baseline without it. While VAPO maintains a near-zero clip ratio, it fails to effectively gate divergent off-policy updates, leading to a rapid training collapse at approximately 90 steps. 

\begin{table*}[ht]
\caption{Ablation results of value model training strategy and critic update frequency. We compare partial parameters, i.e., frozen-attention, with full-parameter value update in RL training, as well as the effectiveness of faster critic updates per policy step. We report Accuracy (\%) for all datasets}
\centering
\setlength{\tabcolsep}{4pt}
\begin{tabular}{lcccc}
\toprule
 & \makecell{Value Training\\Strategy} & \makecell{Critic Update\\Frequency} & AIME2025 & BeyondAIME \\
\midrule
\model & Frozen Attention & 2 & 97.3 & 74.8 \\
\midrule
Single-step-update & Frozen Attention & 1 & 95.00 & 69.75 \\
Full-Parameter Value Training & Full-Parameter & 2 & 90.62 & 74.50 \\
\bottomrule
\end{tabular}
\label{tab:value_training_ablations}
\end{table*}

\begin{table}[ht]
\caption{Ablation results of value model training strategy and critic update frequency. We compare partial parameters, i.e., frozen-attention, with full-parameter value update in RL training, as well as the effectiveness of faster critic updates per policy step. We report Accuracy (\%) for all datasets}
\centering
\begin{tabular}{lcc}
\toprule
  & AIME2025 & BeyondAIME \\
\midrule
\model & 97.3 & 74.8 \\
\midrule
\model w/o Faster value & 95.0 & 69.8 \\
\model w/o Frozen attention  & 90.6 & 74.5 \\
Vanilla VAPO (w/o DIS) & 91.3 & 69.0 \\
Running mean baseline & 79.8 & 55.3 \\
\bottomrule
\end{tabular}
\label{tab:ablations}
\vspace{-3mm}
\end{table}

\subsection{Online Learning Simulation}
\vpara{Task Design.} 
In real-world online learning environments, feedback is typically restricted to a single trajectory per prompt. 
This constraint is inherently incompatible with group-based optimization strategies like GRPO, which depend on relative rewards within a sample group for advantage estimation. In contrast, \model utilizes a value-based critic to provide advantage estimation, allowing for effective policy updates from individual trajectories. 

Therefore, we design a simulated online writing task to assess the adaptability of \model in non-stationary environments. 
In this setting, the feedback signal is designed as the language tone of user preference.
Reward criteria are sequentially adjusted to favor three distinct stylistic archetypes: \textit{cute}, \textit{chuunibyou}, and \textit{classical}. 

\vpara{Dynamic Reward Assignment.} For reward signal assignment, we employ GLM-4.7\citep{2025glm} as an LLM-based judge to evaluate two primary dimensions: response quality and stylistic adherence. The final reward $r \in \{0, 1\}$ is computed as: $r = r_{\text{quality}} \times r_{\text{style}}$ where $r_{\text{quality}}, r_{\text{style}} \in \{0, 1\}$ denote binary rewards. 

Throughout training, the system prompt requires the model to select one stylistic archetype from a pool of four candidates: Academic, Cute, Chuunibyou, and Classical. In our actual experiments, the candidate set consists of \textit{Academic, Cute, Chuunibyou} during the first two phases, and \textit{Classical, Cute, Chuunibyou} in the final phase.

\begin{figure*}[ht]
\centering
\subfloat[We report the accuracy transition of three writing styles---cute, chuunibyou, and classical---on a held-out evaluation set throughout the online training process. Shaded regions indicate phase transitions where the reward preference is switched to favor a different stylistic archetype. \model rapidly suppresses the previously dominant style and realigns its policy to the new target based on environmental feedback.]{
  \includegraphics[width=0.48\textwidth]{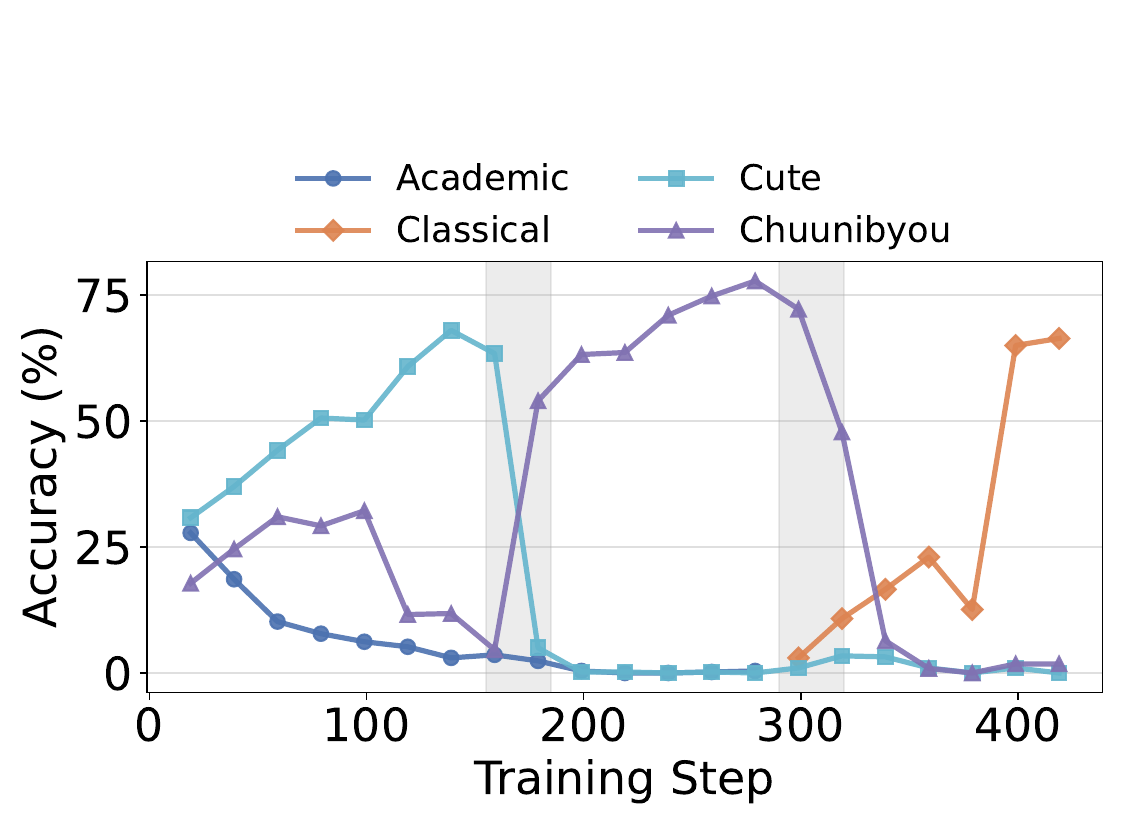}
  \label{fig:writing_eval}
}
\hfill
\subfloat[We compare the evolution of training rewards between \model and a Running Mean Advantage Estimation baseline under single-rollout online learning. Shaded regions denote stylistic reward shifts. While both methods eventually recover after distribution changes, the Running Mean baseline exhibits a pronounced adaptation lag and lower stable performance.]{
  \includegraphics[width=0.48\textwidth]{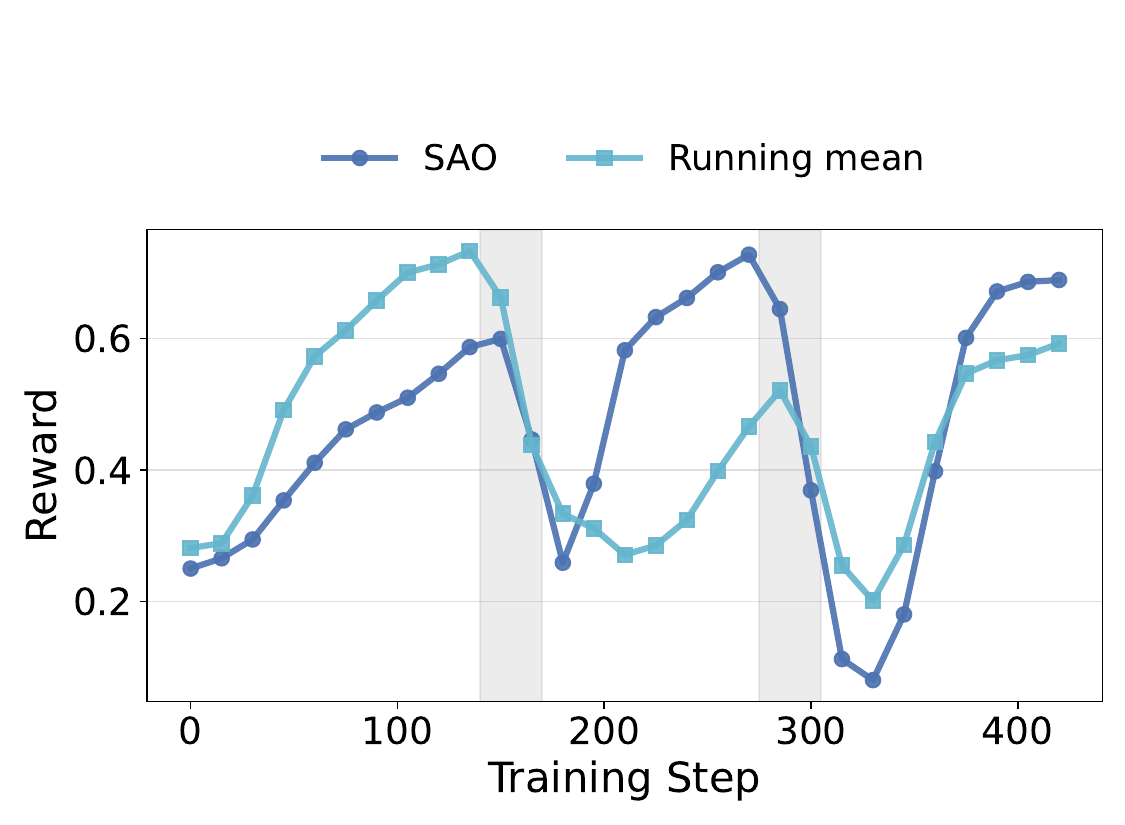}
  \label{fig:writing_training}
}
\caption{Online learning simulation under changing writing-style preferences.}
\label{fig:writing_online}
\end{figure*}

\vpara{Results.} As illustrated in Figure \ref{fig:writing_eval}, we evaluate the performance of the three candidate linguistic styles of each phase on a held-out test set
. \model demonstrates rapid policy realignment following each reward preference shift, characterized by transitions between stylistic archetypes to maintain adherence to the evolving environmental feedback.

\vpara{Comparison against Running-Mean Baseline.} 
To better understand the effectiveness of the value model in the online environment, we also adopt the Running Mean Advantage Estimation approach as the baseline. This method approximates the baseline $b$ by tracking a sliding window of the 128 most recent rewards, thereby facilitating advantage computation as $\hat{A} = r - \mathbb{E}[r_{window}]$. By decoupling advantage estimation from intra-prompt sample groups, this setup permits policy optimization in an online, single-rollout context.
Figure \ref{fig:writing_training} depicts the evolution of training rewards throughout the online learning process of \model and the Running Mean baseline, where the speed and magnitude of reward recovery following stylistic shifts serve as key indicators of algorithmic adaptability. 
The Running Mean baseline exhibits a pronounced adaptation lag due to the inertia of its historical window, which remains temporarily biased by rewards from the preceding distribution. In contrast, \model's value-based critic dynamically tracks reward shifts, facilitating rapid recovery and consistently higher convergence levels. This confirms that \model's state-dependent baseline provides the precision necessary for effective alignment in non-stationary environments.

%% file: 3.related_work.tex
\section{Related Work}

\subsection{Reinforcement Learning for Language Models}
The standard RLHF pipeline trains a reward model from preference data and optimizes the policy with PPO\citep{ouyang2022training, schulman2017proximal}. To reduce the overhead and instability of value-function learning, critic-free objectives such as Group Relative Policy Optimization (GRPO) \citep{deepseek2024math, deepseek2024r1} and REINFORCE-style baselines (e.g., RLOO) \citep{ahmadian2024back} have become increasingly popular. GRPO forms advantages by normalizing rewards within a prompt-level group, which improves stability in synchronous training but introduces an implicit synchronization barrier: updates must wait until all group members are generated, exacerbating staleness and off-policy drift under asynchrony.

Recent work further refines GRPO/PPO-style objectives to improve stability and variance reduction, including sequence-level importance weighting \citep{zheng2025group}, adaptive clipping strategies \citep{yang2025dcpo}, and smoother alternatives to hard clipping \citep{yue2025vapo}. However, these works focus primarily on synchronous RL, where exact importance-sampling ratios are easier to obtain. Importance sampling and clipping strategies for asynchronous RL remain less explored.

\subsection{Synchronous and Asynchronous RL for LLMs}
Most large-scale LLM RL implementations remain synchronous and interleaved: collect a full batch of rollouts with a fixed policy snapshot, then run optimization epochs on that batch \citep{ouyang2022training}. With long-tail output lengths in reasoning and tool-use, synchronous barriers cause stragglers and substantial idle time, motivating asynchronous actor--learner designs where rollout generation and learning proceed concurrently \citep{mnih2016asynchronous, sutton2018reinforcement}. However, asynchrony introduces policy lag and off-policy drift, often requiring staleness-aware training or off-policy corrections \citep{espeholt2018impala}.

Several recent systems target asynchronous RL specifically for LLMs. \citet{noukhovitch2024asynchronous} study asynchronous RLHF as online-but-off-policy learning and characterize robustness tradeoffs. On the systems side, AReaL~\citep{fu2025areal} fully decouples rollout from training and incorporates staleness-aware PPO-style updates for reasoning tasks. ROLL Flash provides fine-grained parallelism and rollout--train decoupling for RLVR and agentic training \citep{lu2025part}.
Complementary to asynchronous systems, MobileRL studies online agentic RL for mobile GUI agents and introduces difficulty-adaptive GRPO variants to improve stability and sample efficiency in multi-turn GUI environments~\citep{xu2025mobilerl}.
Our work complements these systems by focusing on the single-rollout setting where group-based baselines (e.g., GRPO) are structurally mismatched, and by stabilizing asynchronous learning algorithm designs.

%% file: 6.conclusion.tex



\section{Conclusion}

In this work, we explore the optimization of asynchronous RL on the training effectiveness and stability
We proposed \model, a single-rollout asynchronous RL strategy that addresses off-policy and instability. \model stabilizes training with token-level importance sampling and double-sided clipping/masking, and improves generalization by replacing group-wise sampling with single-rollout enabled by stronger value-model training. On agentic reasoning and coding tasks, \model shows consistent outperformance over GRPO baselines, and adapts effectively in simulated online learning.

%% file: 7.appendix.tex
\section{Additional Experimental Results}

\subsection{RL with Agentic Step as Action}
To mitigate the high variance inherent in token-level value predictions, we implement a step-wise GAE calculation. We define a step (denoted as $S_i$) as a single conversation turn and assume that all constituent tokens share a uniform learning signal within each step. We define the step-level value $V(S_i)$ using two primary aggregation methods based on constituent token values predictions $\{v_{i,1}, v_{i,2}, \dots, v_{i,n}\}$: 

\begin{itemize}[leftmargin=*,itemsep=0pt,parsep=0.2em,topsep=0.3em,partopsep=0.3em]
\item \textit{Step Average.} The step value is the average of all token value predictions within that step, $V(S_i) = \frac{1}{n} \sum_{j=1}^{n} v_{i,j}$. In this setting, the value model is trained on all tokens. 
\item \textit{Last-Token Prediction.} We define the step value as $V(S_i) = v_{i,n}$, using only the final token of each step, assuming that the final token provides the most accurate value prediction of the step, as it encapsulates the most comprehensive information of the entire unit. During value model training, we apply a loss mask to all intermediate tokens, ensuring that only the last token of each step contributes to the optimization.
\end{itemize}

To provide a more stable learning signal, we implement a step-wise GAE calculation that shifts the advantage estimation from a token-level to a step-level granularity. We derive a single advantage $\hat{A}_i$ for each step based on the step-level TD error $\delta_i = R_i + \gamma V(S_{i+1}) - V(S_i)$. This advantage is then assigned uniformly to all tokens within the corresponding step, effectively smoothing out the local noise prevalent in auto-regressive generation. Furthermore, the length-adaptive GAE mechanism is modified to scale the decay factor $\lambda$ based on the total number of steps rather than the raw token length, where $$\lambda_{\text{policy}} = 1 - \frac{1}{\alpha * \text{step number}}$$

\begin{figure}[ht]
\centering
\includegraphics[width=0.45\textwidth]{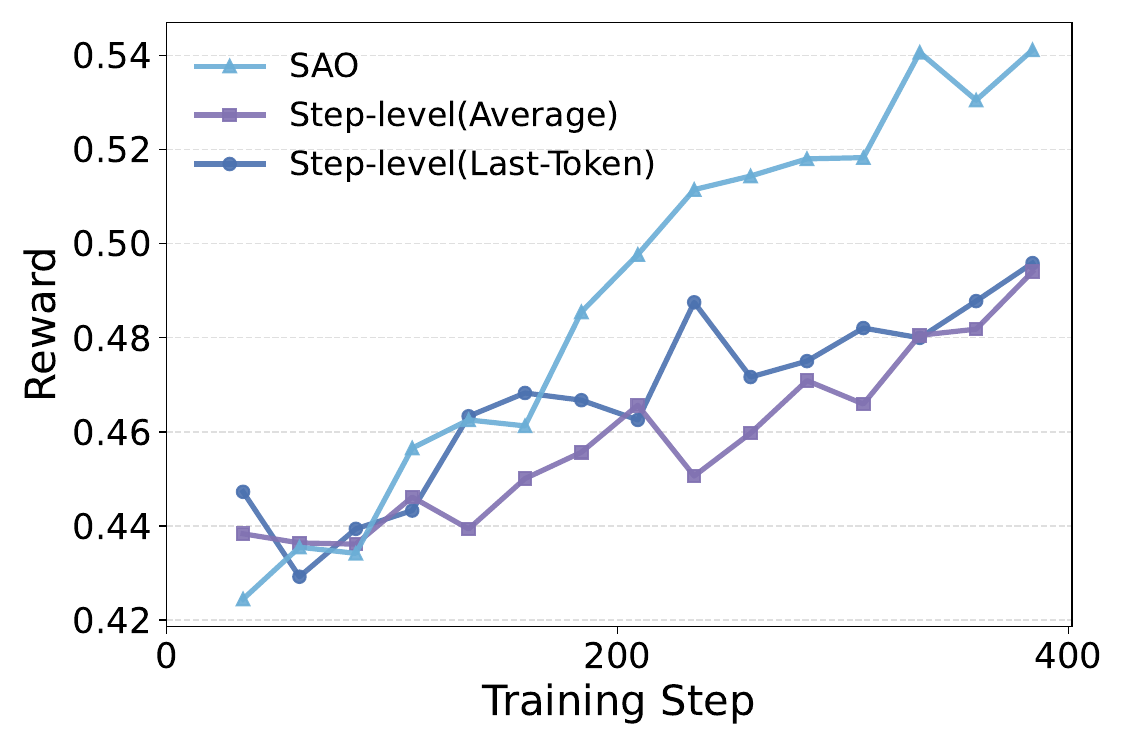}
\caption{Training reward for token-level \model training and step-level variants, where token-level shows better training rewards.}
\label{fig:reward_curve}
\end{figure}

However, as illustrated by the training reward in Figure \ref{fig:reward_curve} and the performance results in Table \ref{tab:stepwise_training}, both step-wise approaches underperform token-wise value training. We attribute this failure to the fact that token-level training provides a finer-grained supervision signal for both the critic and the policy, which is essential for accurately capturing the logical transitions within complex reasoning trajectories.

\begin{table}[htbp]
\centering
\caption{The ablation on the action granularity for value and policy model training. \textit{Step-level} denotes that each agent \textit{step} is viewed as an action to calculate the value. \textit{Token-level} refers to each token being viewed as an action. We report the results with the same training steps (400 steps).}
\setlength{\tabcolsep}{2pt}
\begin{tabular}{@{}lcc@{}}
\toprule
  & AIME2025 & BeyondAIME \\
\midrule
Step-level (Average) & 85.8 & 60.5 \\
Step-level (Last-Token) & 87.3 & 62.8 \\
Token-level & 89.8 & 66.8 \\
\bottomrule
\end{tabular}
\label{tab:stepwise_training}
\end{table}

\subsection{Comparison to Other Baselines of Single-Rollout Strategies}

SPO~\cite{xu2025single} or directly using the historical running-mean reward as the baseline for advantage estimation are also feasible ways to achieve RL with a single rollout per prompt. However, SPO and running-mean baselines rely on prior information about training-data difficulty and achieve worse performance than \model, as shown in the experiment section.

\section{Limitations and Broader Impact}

Our experiments focus on large-scale agentic reasoning, coding, and simulated online writing tasks with a Qwen3-30B-A3B backbone. The conclusions therefore may not transfer directly to smaller models, non-agentic RLHF settings, or environments with dense rewards and shorter rollouts. In addition, \model depends on a trained value model and rollout log-probabilities, so deployment requires infrastructure that can reliably preserve token-level behavior probabilities during asynchronous generation. The online learning study uses a controlled simulated preference shift; real user-facing online adaptation would require stronger safeguards, monitoring, and privacy review before deployment.

By improving the stability and efficiency of LLM reinforcement learning, this work can reduce the cost of training capable agentic systems. The same capability could also make it easier to optimize models for harmful objectives if used without appropriate data filtering, access controls, or evaluation, so responsible release and monitoring are important for any deployed system derived from this work.